\documentclass[letterpaper, 10 pt, conference]{ieeeconf}  

\usepackage{amsmath,graphicx}
\usepackage{float}
\usepackage{subfigure}
\usepackage{multirow}

\IEEEoverridecommandlockouts                              

\title{\LARGE \bf
Multi-Cell Multi-Task Convolutional Neural Networks\\ for Diabetic Retinopathy Grading
}

\author{Kang Zhou$^{1,2}$,  Zaiwang Gu$^{2,3}$, Wen Liu$^{1}$, Weixin Luo$^{1}$, Jun Cheng$^{2}$, Shenghua Gao$^{1}$, Jiang Liu$^{2}$
\thanks{$^1$School of Information Science and Technology, ShanghaiTech University, China. {\tt\small \{zhoukang, liuwen, luowx, gaoshh\}  @shanghaitech.edu.cn}}
\thanks{$^2$Ningbo Institute of Materials Technology and Engineering, China. {\tt\small \{guzaiwang, chengjun, jimmyliu\}@nimte.  ac.cn}}
\thanks{$^3$Shanghai University, China.}    	
}

\begin{document}

\maketitle
\thispagestyle{empty}
\pagestyle{empty}

\begin{abstract}
	Diabetic Retinopathy (DR) is a non-negligible eye disease among patients with Diabetes Mellitus, and automatic retinal image analysis algorithm for the DR screening is in high demand. Considering the resolution of retinal image is very high, where small pathological tissues can be detected only with large resolution image and large local receptive field are required to identify those late stage disease, but directly training a neural network with very deep architecture and high resolution image is both time computational expensive and difficult because of gradient vanishing/exploding problem, we propose a \textbf{Multi-Cell} architecture which gradually increases the depth of deep neural network and the resolution of input image, which both boosts the training time but also improves the classification accuracy. Further, considering the different stages of DR actually progress gradually, which means the labels of different stages are related. To considering the relationships of images with different stages, we propose a \textbf{Multi-Task} learning strategy which predicts the label with both classification and regression. Experimental results on the Kaggle dataset show that our method achieves a Kappa of 0.841 on test set which is the 4-th rank of all state-of-the-arts methods. Further, our Multi-Cell Multi-Task Convolutional Neural Networks (M$^2$CNN) solution is a general framework, which can be readily integrated with many other deep neural network architectures.
	\par \emph{Index Term---}Deep Learning, Multi-Cell Architecture, Multi-Task Learning, Medical Image, Diabetic Retinopathy.

\end{abstract}

\section{INTRODUCTION} 
\begin{figure*}[t]
	\centering
	\subfigure[Grade 0]{
		\label{fig:subfig:a} 
		\includegraphics[width=1.0in, height = 0.7in]{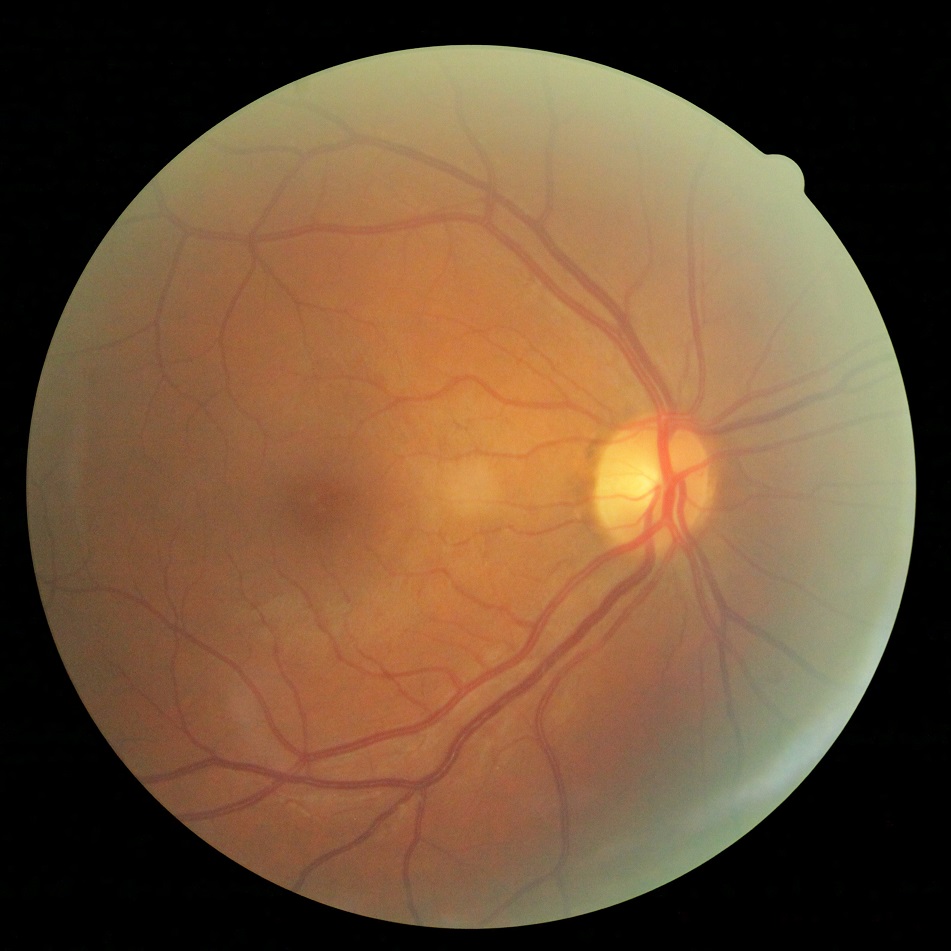}}
	\subfigure[Grade 1]{
		\label{fig:subfig:b} 
		\includegraphics[width=1.0in, height = 0.7in]{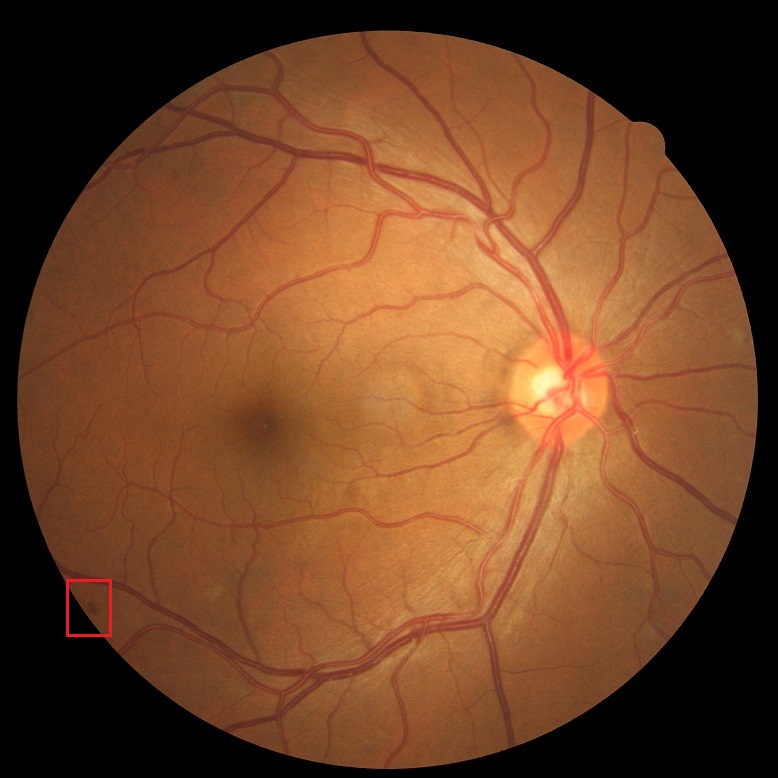}}
	\subfigure[Grade 2]{
		\label{fig:subfig:c} 
		\includegraphics[width=1.0in, height = 0.7in]{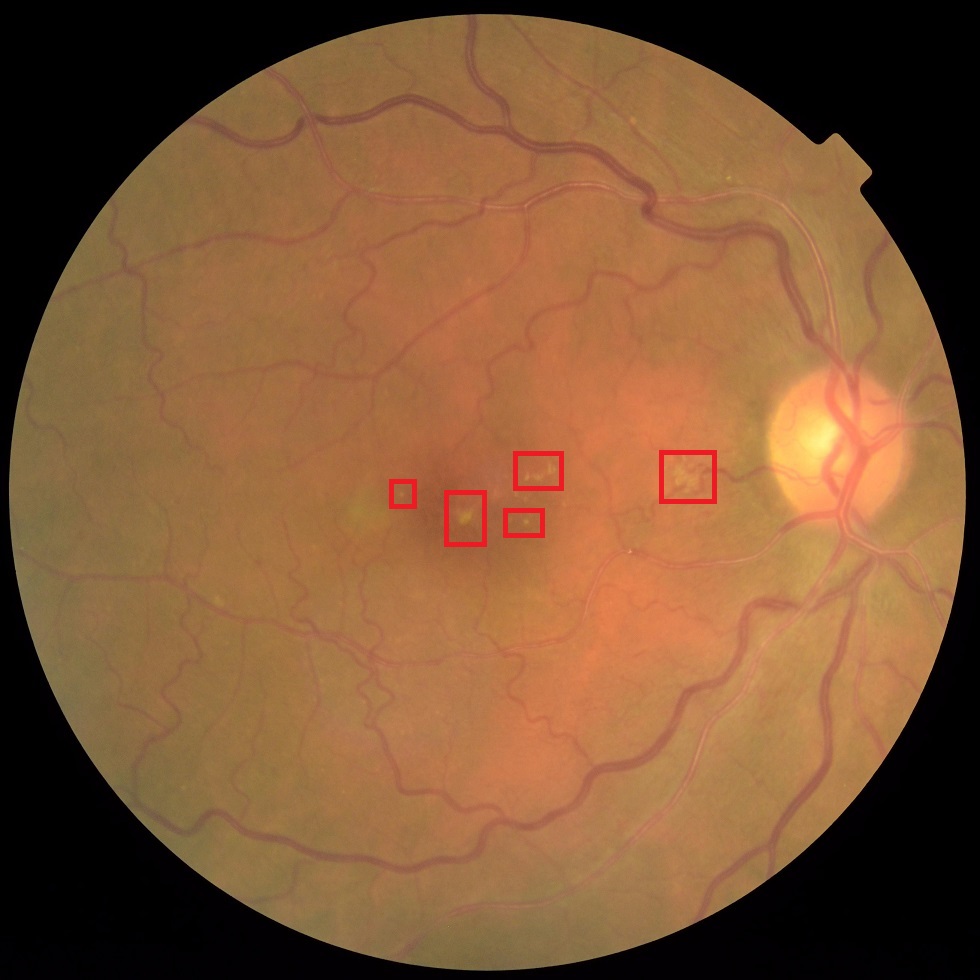}}
	\subfigure[Grade 3]{
		\label{fig:subfig:d} 
		\includegraphics[width=1.0in, height = 0.7in]{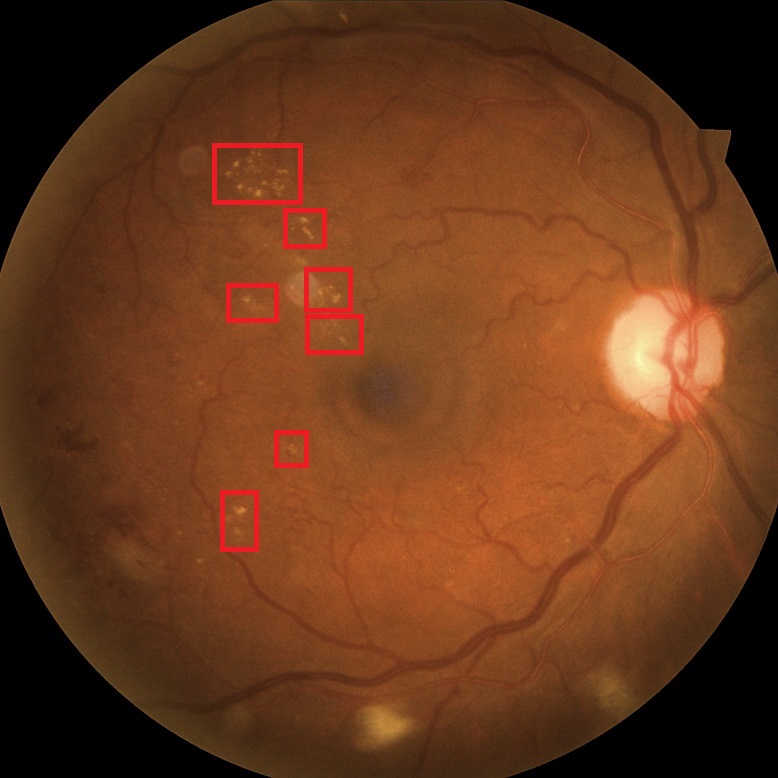}}
	\subfigure[Grade 4]{
		\label{fig:subfig:e} 
		\includegraphics[width=1.0in, height = 0.7in]{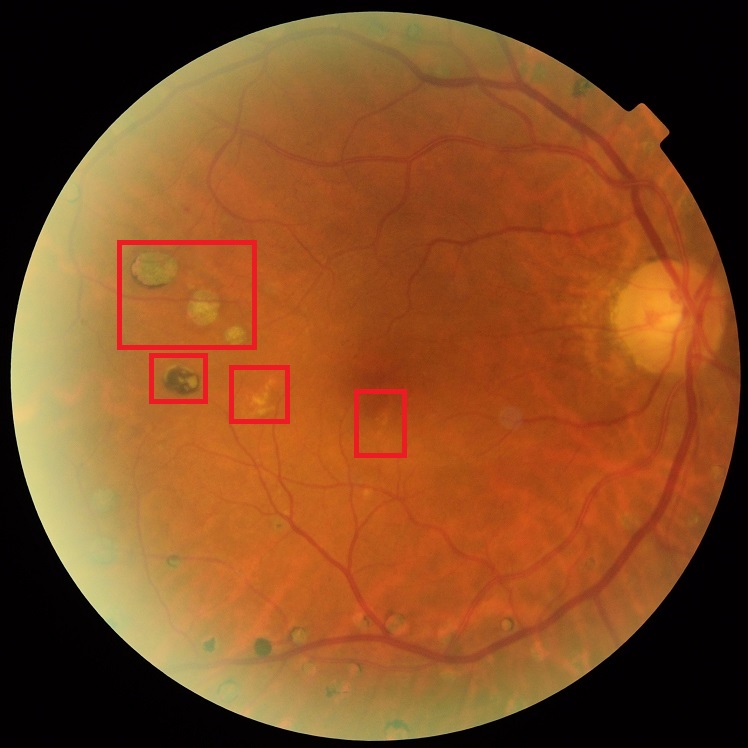}}
	\caption{5 grades DR image instances. Grade 0, 1, 2, 3, 4 means absence of DR, mild non-proliferative DR (NPDR), moderate non-proliferative DR, sever non-proliferative DR, and proliferative DR(PDR), respectively.}
	\label{figure:5class} 
\vspace{-0.7cm}
\end{figure*}

Diabetic Retinopathy (DR) is an eye disease caused by diabetes. Usually DR distresses people who has diabetes for a significant number of years. It will lead DR patients to blindness if untreated while treatments can be applied to slow down or stop further vision loss if the condition can be detected early. It is of great significance for people with diabetes to have a regular eye screening for DR. Clinicians often use a five-grade system as shown in Fig. \ref{figure:5class} to describe the severity of DR. Currently the image grade is obtained manually, which is time-consuming, subjective and expensive. Therefore, automatic retinal image analysis algorithm for DR grading is in high demand.

\par Most automated systems \cite{cree1997fully}\cite{fleming2006automated}\cite{venkatesan2012classification} use hand-crafted image features, such as shape, color, brightness and domain knowledge of diabetic retinopathy, which includes optic disk, blood vessels and macula. Since 2012, Convolution Neural Networks (CNN) have shown remarkable performance in image classification tasks \cite{krizhevsky2012imagenet}. Recently, there are some works for DR diagnosis at image level based on deep learning. Machael \emph{et al}. \cite{abramoff2016improved} compared the performance between deep learning method and previous traditional methods, and shows that deep learning achieves significantly better performance in DR diagnosis. Chanrakumar \emph{et al}. \cite{naduclassifying} used Deep CNN to classify fundus images on the public Kaggle \cite{kaggle.com} dataset with image-level labels.

\par However, because of the characteristics of DR images, deep learning based DR diagnosis has two challenges: i) The image resolution of DR images (usually 2048 $\times$ 3072 pixels or larger) is significantly higher than that of general images (469 $\times$ 387 pixels on ImageNet benchmark \cite{deng2009imagenet}). On the one hand, such high resolution is required because those small pathological tissues can be found only with high resolution images. So directly reducing the resolution of DR images with downsampling as the input of CNN would evidently reduces the sensitivity of CNN to these early stage disease. But the network training with such large resolution image is very time consuming. On the other hand, for those late stage disease, the large local receptive field is needed to identify the disease with larger regions. To increase the local receptive field, we can either reduce the image resolution for the fixed depth CNN, which is not desirable for early stage disease, or increase the kernel size or depth for the fixed resolution of image, which may lead to gradient vanishing/exploding problem and more expensive computational costs because of more parameters. To tackle this problem, we design a Multi-Cell architecture. We gradually increase the resolution of images and the depth of CNN which not only accelerates the training procedure but also improves the diseases classification accuracy. ii) For general image classification, if one image is misclassified, it's loss is fixed and it is not related to which category the image is classified to. However, for DR diagnosis, on the one hand, we want the image to be corrected; on the other hand, the diseases progress gradually, so the severities of diseases at different stages are different, so the loss of misclassifying diseases to different incorrect stages are different, either. For example, the price of misclassifying a proliferative DR (grade 4) as no DR (grade 0) is much higher than that of misclassifying a mild non-proliferative DR as no DR. In other words, even if the image is not classified, we want it's predicted label so to as close to the ground truth as possible. So both the popular softmax loss (or Cross Entropy, CE) in classification and Mean Square Error (MSE) loss in regression for general computer vision tasks are not optimal for medical imaging. Thus we propose a Multi-Task Learning strategy: we use not only CE loss for image classification to guarantee the correctness of DR diagnosis but also MSE loss for regression to guarantee the small discrepancy between the ground-truth and predicted label. We term our solution as Multi-Cell Multi-Task Convolutional Neural Networks (M$^2$CNN) based DR grading.

\par The contributions of our work can be summarized as follows: i) We propose a Multi-Cell CNN architecture which not only accelerates the training procedure, but also improves the classification accuracy; ii) We propose a Multi-Task Learning strategy to simultaneously improves the classification accuracy and discrepancy between ground-truth and predicted label; iii) Experimental results validate the effectiveness of our method. Further, our solution can be readily integrated with many other existing CNN based DR image diagnosis and other disease diagnosis.

\begin{figure}[t]
	\begin{center}
	\includegraphics[width=0.75\linewidth,height = 12cm]{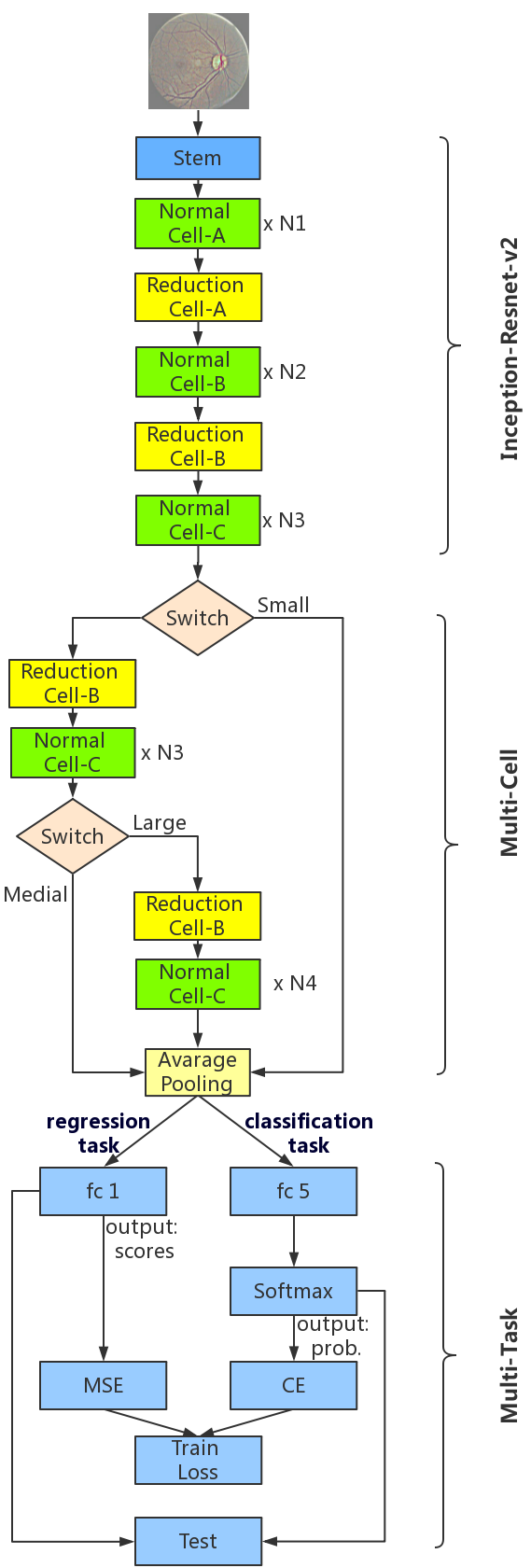}
	\end{center}
	\caption{The overall network architecture of our M$^2$CNN. The \textbf{stem} proposed in Inception-Resnet-v2 \cite{szegedy2017inception} consists of some convolutional layers and max pooling layers. ``Cell'' means convolutional layers and regard these layers as a whole. The two types of cell are \textbf{normal cell} that returns a feature map of the same dimension and \textbf{reduction cell} that returns a feature map where the height and width are reduced. \textbf{Multi-Cell} architecture chooses the neural network path based on the resolution of input images. \textbf{Training loss} includes both Mean Square Error (MSE) loss and Cross Entropy (CE) loss. In \textbf{testing} phase, we can either use scores and probabilities to predict the label, and our experiments on Kaggle show that scores based prediction usually achieves better performance.}
	\label{fig:overall}
\vspace{-0.6cm}
\end{figure}

\begin{figure}[t]
	\centering
	\subfigure[Original image]{
		\label{fig:subfig:a} 
		\includegraphics[height=0.8in, width=1.4in]{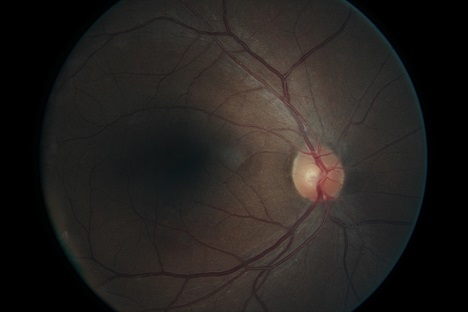}}
	\subfigure[Preprocessed image]{
		\label{fig:subfig:b} 
		\includegraphics[height=0.8in,  width=1.1in]{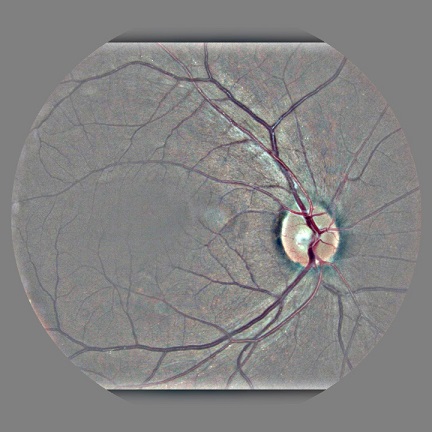}}
	\caption{Image Preprocessing}
	\label{figure:preprocess} 
\vspace{-0.4cm}
\end{figure}
\section{Our Method} \label{method}


The overall architecture of our M$^2$CNN is shown in Fig.{\ref{fig:overall}}. It consists three modules: 1) Inception-Resnet-v2 module \cite{szegedy2017inception}. It is used as BaseNet; 2) Multi-Cell architecture module. It chooses different neural network path depending on the resolution of input medical image, and gradually increases the depth and local receptive field for higher resolution images. We use small and medial resolution images to pre-train the model to accelerate the train speed for large resolution images; 3) Multi-Task Learning module. We simultaneously conduct image classification and label regression task. 


\subsection{Preprocessing}
Since the fundus retinal images are captured under different conditions, these images often vary largely from one to another in terms of lighting condition, color, the ratio of the fundus area in the image, \emph{etc}. In this paper, we apply the following preprocessing to reduce the variation. We firstly remove the extra black pixels from the image and then perform  the image normalization based on the min-pooling filtering \cite{graham2015kaggle}.
\begin{equation}
	I_c = \alpha I + \beta G(\rho)*I + \gamma
\end{equation}
Here $*$ denotes the convolution operation, $I$ denotes input image and $G(\rho)$ represents the Gaussian filter with a standard deviation of $\rho$. Fig.{\ref{figure:preprocess}} shows an example of the original image and its associated preprocessed one.


\subsection{Multi-Task Learning}
A commonly used loss function for classification in general computer vision is CE loss function. It outputs probabilistic predictions by using softmax activation. The loss function of CE is
\begin{equation}
	\label{equation:ce}
	L_1 = -\frac{1}{m}\bigg[ \sum^m_{i=1} \sum^k_{j=1} 1\{y^{(i)}=j\}\log(\text{Prob}_j)\bigg]
\end{equation}
where $m$ denotes the number of input instances, $1 \{ \cdot \}$ denotes the indicator function, $y^{(i)}$ denotes the $i$-th label and $\text{Prob}_j$ denotes the probabilities output by softmax activation. However, if some fundus is misclassified, it doesn't consider the difference between that different stages that image is classified into. However, as we aforementioned, the price of misclassifying a proliferative DR (grade 4) as no DR (grade 0) is much higher than that of misclassifying a mild non-proliferative DR as no DR. In other words, even if the image is not classified, we want its predicted label so to as close to the ground truth as possible. To achieve this goal, we can leverage the Mean Square Error (MSE) in the regression task, which is defined as follows:
\begin{equation}
	\label{equation:mse}
	L_2 = \frac{1}{m}\sum_{i=1}^m(y-y^{(i)})^2
\end{equation}
where $y$ is the output score. Although MSE considers the distance between a false prediction and the true label, there is an issue that it is usually more difficult to optimize than that of classification. For example, if the distance $\delta = |y - y^{(i)}|$ is small the squared value $\delta ^2$ will be smaller and it will be too small to optimize.

\par In this paper, we propose to integrate the MSE loss with CE loss. Since MSE computes the distance between different classes, it complements the CE loss. The proposed loss function is defined as follows:
\begin{equation}
	L = L_1 + L_2 + L_{reg}
\end{equation}
where $L_{reg}$ denotes the regularization loss (weight decay term) used to avoid overfitting.


\subsection{Multi-Cell Architecture}

Since the downsampling of the original retinal image with large resolution often leads to information loss especially when the lesion is small, it is not optimal to down sample the image into a very small size, e.g. 224 $\times$ 224 pixels, that is often used in general computer vision. On the other side, if the input image is large and we pass it to BaseNet architectures, it will introduce more computational costs. Further, for late stage disease, the large local receptive field is needed, which would cause the increase of kernel size/depth, which may lead to the gradient vanishing/exploding problem in optimization. To facilitate the training of CNN with large resolution image, we propose Multi-Cell architecture by gradually increasing the resolution of the image and the depth of network.
%

\begin{table}[htp]
	\newcommand{\tabincell}[2]{\begin{tabular}{@{}#1@{}}#2\end{tabular}} 
	\linespread{0.8}
	\normalsize
	\caption{Spatial Resolution of Input Image and some Feature Map}
	\begin{center}
		\setlength{\tabcolsep}{0.8mm}
		\begin{tabular}{c|c|c|c|c}\hline
			\tabincell{c}{input image} & 224$\times$224 & 256$\times$256 & 448$\times$448 & 720$\times$720\\ \hline
			\tabincell{c}{before switch}  & 5$\times$5 & 8$\times$8 & 12$\times$12 & 21$\times$21 \\ \hline
			\tabincell{c}{after multi-cell} & 5$\times$5 & 8$\times$8 & 5$\times$5 & 4$\times$4 \\ \hline
		\end{tabular}
	\end{center}
	\label{label:multi_size}
\vspace{-0.6cm}
\end{table}

\par As shown in Fig.{\ref{fig:overall}}, the multi-cell architecture module chooses the convolutional neural network path (depth of network) based on the resolution of input image: i) When the input image is small(resolution $<$ 350 $\times$ 350 pixels), the output before switch goes directly to average pooling; ii) When the input image is medium size (350 $\times$ 350 pixels $<=$ resolution $<$ 700 $\times$ 700 pixels), it will go through the Reduction Cell-B and $N_3$ Normal Cell-C (Medial Image Cells) before arriving average pooling. Reduction Cell-B and Normal Cell-C have same convolutional architectures as them in BaseNet ; 3) When it is large (700 $\times$ 700 pixels $<=$ resolution), it will go through the Medial Image Cells and another Reduction Cell-B, $N_4$ Normal Cell-C before arriving average pooling. We use the small resolution image to pre-train the network BaseNet, and use medium resolution image to pre-train the model of BaseNet and Medial Image Cells to accelerate the training speed with large resolution images.  

The advantages of such solution can be summarized as follows: i) It accelerates the network training, and such pre-training facilitates the parameter optimization in the training phase, consequently improves the classification accuracy; ii) We use large resolution image as input which improves the sensitivity for the detection of early stage diseases; iii) We use deeper network to increase the local receptive field which helps the detection of late stage diseases.


\section{EXPERIMENTS} \label{results}

\subsection{Experimental Setup}
\textbf{Dataset.} Kaggle organized a comprehensive competition in order to design an automated retinal image diagnosis system for DR screening in 2015 \cite{kaggle.com}. The retinal images were provided by EyePACS, which is a free platform for retinopathy screening. The dataset consists of 35126 training images, 10906 validate images and 42670 test images. Each image is labeled as $\{0,1,2,3,4\}$ and the number represents the level of DR. Following the work \cite{wang2017zoominnet},  we also use the Kaggle dataset to evaluate our algorithm.

\textbf{Evaluation Metric.} We use the quadratic weighted kappa to evaluate our proposed methods, which is used in Kaggle DR Challenge. The quadratic weighted kappa not only measures the agreement between two ratings but also considers the distance between the prediction and the ground truth. 

\textbf{Parameter Initialization.} Following the work \cite{wang2017zoominnet}, in our experiments, for BaseNet, it is initialized with parameters trained for ImageNet classification. For M$^2$CNN, it is trained based on the weights of BaseNet+MT since it consists of BaseNet+MT and Multi-Cell.

\textbf{Hyper-parameters.} During the preprocessing phrase, the value of $\alpha, \beta, \rho, \gamma$ are empirically fixed as $\alpha=4, \beta=-4, \rho=10, \gamma=128$. For the overall algorithm as show in Fig. {\ref{fig:overall}}, we set $N_1=10, N_2=20, N_3=10, N_4=5$.  During the training, our network use fine-tune strategy, so we use two learning rates:  $LR_1$=0.0001 for fine tuning weights and  $LR_2$=0.001 for learning other weights in our M$^2$CNN. We also list the spatial resolution of input image and feature maps in our M$^2$CNN in Table \ref{label:multi_size}.


%
%

\subsection{Evaluation of Different Modules}

\par
\begin{figure}[h]
	\begin{center}
		\includegraphics[width=1\linewidth,height=4cm]{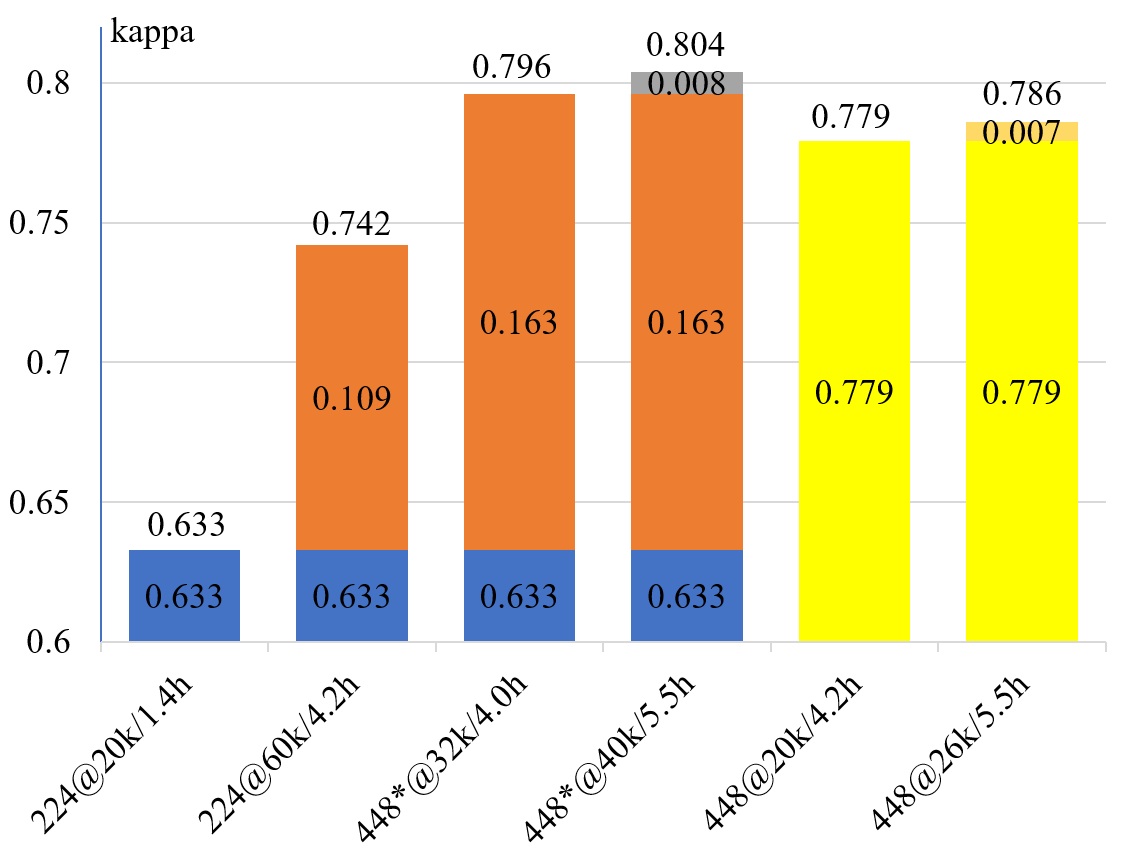}
	\end{center}
	\caption{Comparison of different training method. ``224@20k/1.4h'' denotes the input resolution is 224 $\times$ 224 pixels, the number of training steps is 20,000 and it costs 1.4 hours to train. ``448*@32k/4.0h'' denotes  it is fine-tuned on the model of ``224@20k/1.4h" and train another 12,000 (32,000 - 20,000) steps using  448 $\times$ 448 pixels input images. ``448@20k/4.2h'' denotes it is trained using 448 $\times$ 448 pixels input images without fine-tuning.}
	\label{fig:histogram}
\vspace{-0.5cm}
\end{figure}

 To evaluate each module of M$^2$CNN, we conduct ablation experiments and the results are shown in Table {\ref{label:multi_task}}. All of these results in Table {\ref{label:multi_task}} and Fig. \ref{fig:histogram} are evaluated on validation set.

\begin{table}[htb]
\newcommand{\tabincell}[2]{\begin{tabular}{@{}#1@{}}#2\end{tabular}}
\normalsize
\caption{Results of Each Module}
\begin{center}
	  \setlength{\tabcolsep}{1.2mm}
	  \begin{tabular}{|c|c|c|c|c|c|c|}\hline
		  Train   & MSE & CE & \multicolumn{2}{|c|}{Multi-Task} & \multicolumn{2}{|c|}{ M$^2$CNN } \\ \cline{1-7}
		  Test	  & scores & prob.  & scores & prob. & scores & prob. \\ \hline
		  224$\times$224 & 0.720 & 0.725 & 0.742   & 0.718  & -   & -  \\ \hline
		  448$\times$448 & 0.790 & 0.772 & 0.812   & 0.782  & \textbf{0.830}   & 0.812  \\ \hline		
		  720$\times$720 & 0.835 & 0.751 & \textbf{0.841}   & 0.826  & \textbf{0.844}   & 0.842  \\ \hline
	  \end{tabular}
  \end{center}
  \label{label:multi_task}
  \vspace{-0.5cm}
\end{table}

\par \textbf{\textit{Multi-Task Learning Module:}} The results of the Multi-Task column in Table {\ref{label:multi_task}} show that large image resolution is important for DR diagnosis. No matter using scores or probabilities (prob.) test method, both performances of multi-task learning are better than that of single task. That is to say, when optimizing the multi-task loss function, MSE loss and CE loss help each other to achieve a better local optimum in solution space. Similar results can also be found in Table \ref{label:final_result} by comparing BaseNet with BaseNet+MT on testing set. Further, we find that scores based evaluation usually corresponds to better performance, so we use scores based evaluation.

\par \textbf{\textit{Multi-Cell Architecture Module:}} To verify multi-cell can accelerate the training speed, we conduct experiments by training with large resolution images directly, and the results are shown in Fig. \ref{fig:histogram}. We can see that the result of ``448*@32k/4.0h'' is better than ``224@60k/4.2h'' and ``448@20k/4.2h'', while the result of ``448*@40k/5.5h'' is better than ``448@26k/5.5h''. That is to say, using multi-cell architecture with fine-tune strategy, we can get a better performance. Similar results can also be found in Table \ref{label:final_result} by comparing M$^2$CNN with BaseNet+MT on the testing set. All of these results are based on multi-task learning.

%

\subsection{Performance Comparison} We also compare our M$^2$CNN with the methods achieve the best performance on Kaggle challenge and the state-of-the-art method \cite{wang2017zoominnet} for CNN based DR diagnosis. The results are shown in Table Table {\ref{label:final_result}}. We can see that our M$^2$CNN algorithm achieves 3-rd rank (top 0.45\%) in Kaggle challenge, which validates the effectiveness of our method. Further, our solution is related to the input and network optimization, so it can be readily integrated with other CNN based DR diagnosis network and can be applied to the diagnosis of other diseases.

\begin{table}[h]
\vspace{-0.3cm}
	\normalsize
	\caption{Comparison with other algorithms}
	\begin{center}
		\setlength{\tabcolsep}{4mm}
		\begin{tabular}{l c c} \hline
			Algorithm                            & val set & test set \\ \hline
			Min-pooling                          & 0.860   & 0.849    \\
			Zoom-in-Net	\cite{wang2017zoominnet} & 0.857   & 0.849	\\
			o\_O                                 & 0.854   & 0.844    \\
			Reformed Gamblers                    & 0.851   & 0.839    \\ 
			M-Net+A-Net \cite{wang2017zoominnet} & 0.837   & 0.832    \\ \hline
			BaseNet                              & 0.835   & 0.828        \\
			BaseNet+MT                           & 0.841   & 0.838    \\
			M$^2$CNN	                         & 0.844   & 0.841        \\ \hline
		\end{tabular}
	\end{center}
	\label{label:final_result}
\vspace{-0.7cm} 
\end{table}  
\section{CONCLUSION}
In this paper, based on the characteristics of DR image, we design a novel Multi-Cell Multi-Task Convolutional Neural Networks (M$^2$CNN), which can tackle the DR diagnosis with high resolution images and improves the classification. Experimental results validate the effectiveness of our solution for DR image classification. 
 

\section{Acknowledge}
The project is supported by NSFC (NO. 61502304 ), Shanghai Subject Chief Scientist (A type) (No. 15XD1502900) and grant under Y80002RA01, Y60001RA01, Y61102DL03, Y50709WR08 by Chinese Academy of Sciences.

 {\small
  \bibliographystyle{ieee}
  \bibliography{research}

\begin{thebibliography}{10}\itemsep=-1pt

\bibitem{kaggle.com}
Diabetic retinopathy detection.
\newblock \url{https://www.kaggle.com/c/diabetic-retinopathy-detection/data}.

\bibitem{abramoff2016improved}
M.~D. Abr{\`a}moff, Y.~Lou, A.~Erginay, W.~Clarida, R.~Amelon, J.~C. Folk, and
  M.~Niemeijer.
\newblock Improved automated detection of diabetic retinopathy on a publicly
  available dataset through integration of deep learning.
\newblock {\em Investigative ophthalmology \& visual science},
  57(13):5200--5206, 2016.

\bibitem{cree1997fully}
M.~J. Cree, J.~A. Olson, K.~C. McHardy, P.~F. Sharp, and J.~V. Forrester.
\newblock A fully automated comparative microaneurysm digital detection system.
\newblock {\em Eye}, 11(5):622--628, 1997.

\bibitem{deng2009imagenet}
J.~Deng, W.~Dong, R.~Socher, L.-J. Li, K.~Li, and L.~Fei-Fei.
\newblock Imagenet: A large-scale hierarchical image database.
\newblock In {\em Computer Vision and Pattern Recognition, 2009. CVPR 2009.
  IEEE Conference on}, pages 248--255. IEEE, 2009.

\bibitem{fleming2006automated}
A.~D. Fleming, S.~Philip, K.~A. Goatman, J.~A. Olson, and P.~F. Sharp.
\newblock Automated microaneurysm detection using local contrast normalization
  and local vessel detection.
\newblock {\em IEEE transactions on medical imaging}, 25(9):1223--1232, 2006.

\bibitem{graham2015kaggle}
B.~Graham.
\newblock Kaggle diabetic retinopathy detection competition report.
\newblock {\em University of Warwick}, 2015.

\bibitem{krizhevsky2012imagenet}
A.~Krizhevsky, I.~Sutskever, and G.~E. Hinton.
\newblock Imagenet classification with deep convolutional neural networks.
\newblock In {\em Advances in neural information processing systems}, pages
  1097--1105, 2012.

\bibitem{naduclassifying}
T.~Nadu.
\newblock Classifying diabetic retinopathy using deep learning architecture.

\bibitem{szegedy2017inception}
C.~Szegedy, S.~Ioffe, V.~Vanhoucke, and A.~A. Alemi.
\newblock Inception-v4, inception-resnet and the impact of residual connections
  on learning.
\newblock In {\em AAAI}, pages 4278--4284, 2017.

\bibitem{venkatesan2012classification}
R.~Venkatesan, P.~Chandakkar, B.~Li, and H.~K. Li.
\newblock Classification of diabetic retinopathy images using multi-class
  multiple-instance learning based on color correlogram features.
\newblock In {\em Engineering in Medicine and Biology Society (EMBC), 2012
  Annual International Conference of the IEEE}, pages 1462--1465. IEEE, 2012.

\bibitem{wang2017zoominnet}
Z.~Wang, Y.~Yin, J.~Shi, W.~Fang, H.~Li, and X.~Wang.
\newblock Zoom-in-net: Deep mining lesions for diabetic etinopathy detection.
\newblock In {\em MICCAI}, 2017.

\end{thebibliography}
 }

\end{document}